%% file: samplepaper.tex
\documentclass[runningheads]{llncs}
\usepackage[T1]{fontenc}
\usepackage{graphicx}

\usepackage{amssymb}
\usepackage{soul}
\usepackage{url}
\usepackage{ulem}
\usepackage{booktabs}
\usepackage{amsmath}

\usepackage{amsfonts}  
\usepackage{amssymb}
\usepackage{multicol}
\usepackage{multirow}
\usepackage{graphicx}
\usepackage{booktabs}
\usepackage{colortbl}
\usepackage{stfloats}
\usepackage{comment}
\usepackage{makecell}
\usepackage[noend]{algpseudocode}
\usepackage{algorithmicx,algorithm}
\usepackage{pifont}
\usepackage{color}
\usepackage{microtype}
\usepackage{cleveref}

\begin{document}
\title{Co-evolving Graph Reasoning Network for Emotion-Cause Pair Extraction}
%
%
\author{Bowen Xing\inst{1,2}\and
Ivor W. Tsang \inst{2,1}}
\authorrunning{Bowen \& Ivor.}
%
\institute{Australian Artificial Intelligence Institute, University of Technology Sydney\\
 \and
Centre for Frontier AI Research, A*STAR\\
\email{bwxing714@gmail.com  \quad ivor\_tsang@cfar.a-star.edu.sg}}
\maketitle              
\begin{abstract}Emotion-Cause Pair Extraction (ECPE) aims to extract all emotion clauses and their corresponding cause clauses from a document. Existing approaches tackle this task through multi-task learning (MTL) framework in which the two subtasks provide indicative clues for ECPE. However, the previous MTL framework considers only one round of multi-task reasoning and ignores the reverse feedbacks from ECPE to the subtasks. Besides, its multi-task reasoning only relies on semantics-level interactions, which cannot capture the explicit dependencies, and both the encoder sharing and multi-task hidden states concatenations can hardly capture the causalities. To solve these issues, we first put forward a new MTL framework based on Co-evolving Reasoning. It (1) models the bidirectional feedbacks between ECPE and its subtasks; (2) allows the three tasks to evolve together and prompt each other recurrently; (3) integrates prediction-level interactions to capture explicit dependencies. Then we propose a novel multi-task relational graph (MRG) to sufficiently exploit the causal relations. Finally, we propose a Co-evolving Graph Reasoning Network (CGR-Net) that implements our MTL framework and conducts Co-evolving Reasoning on MRG. Experimental results show that our model achieves new state-of-the-art performance, and further analysis confirms the advantages of our method.

\keywords{Multi-Task Learning \and  Relational Graph Reasoning \and Emotion-Cause Extraction \and Natural Language Processing.}
\end{abstract}
%
\input{introduction.tex}

\input{relatedwork.tex}
\input{method.tex}
\input{experiment.tex}

\input{conclusion.tex}

%
%
%
\bibliographystyle{splncs04}
\bibliography{anthology.bib}

\end{document}

%% file: introduction.tex
\section{Introduction}\label{sec:introduction}
\begin{figure}[t]
 \centering
 \includegraphics[width = 0.4\textwidth]{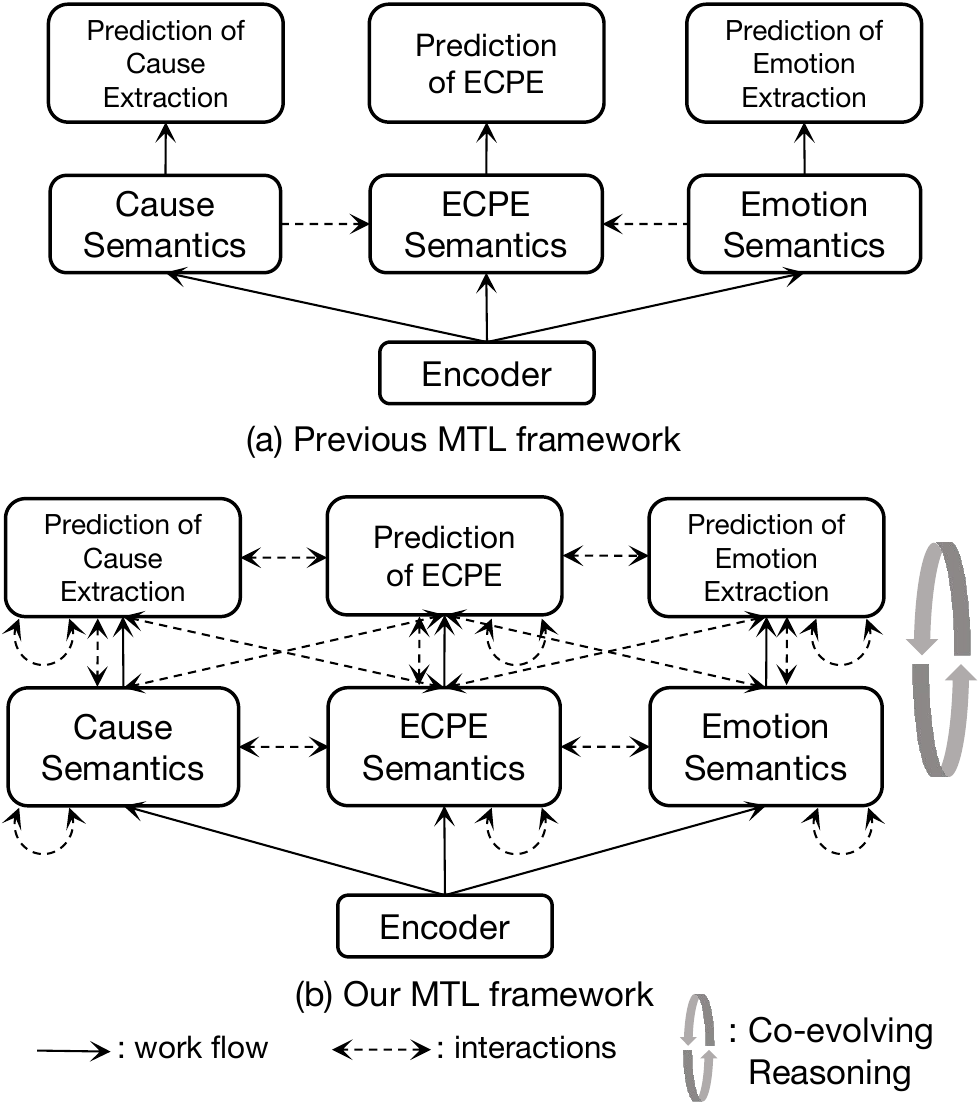}
 \caption{Comparison of the previous MTL framework and our MTL framework for ECPE.}
 \label{fig: framework}
\end{figure}

Emotion-Cause Pair Extraction (ECPE) is a new while challenging task in the field of natural language processing/artificial intelligence. 
It aims to automatically extract all emotion clauses and the corresponding cause clauses from a raw document, which is of great value for real-world application \cite{ecpe}.
Consider a document ``$[$\textit{In the memory of the students}$]_1$, $[$\textit{he often paid the tuition fees for them}$]_2$, $[$\textit{which is respectable and touching}$]_3$.'', the third clause expresses an emotion, which is triggered by the second clause, so these two clauses form an emotion-cause pair. 
Intuitively, detecting the clauses that express causes and emotions, namely cause extraction (CE) and emotion extraction (EE), are two subtasks of ECPE.
Accordingly, recent models \cite{ecpe2d,mtst,mgsag,maca} implement the multi-task learning framework shown in Fig.\ref{fig: framework}(a), introducing CE and EE to provide indicative clues for ECPE.

Although the previous multi-task learning (MTL) framework has made promising progress, based on our observation, it still suffers from several issues which hinder the multi-task reasoning between ECPE and CE/EE.
First, previous works only consider one-way messages from CE/EE to ECPE. 
The predictions of CE/EE may be incorrect due to their unreliable semantics.
In this case, the false information from CE/EE may mislead ECPE.
Second, the single-round multi-task reasoning process in previous works is not competent, considering that it is hard for machines to understand emotions, causes, and their causalities like humans due to the inherent ambiguity and subtlety of emotions and causes \cite{transecpe}.
Third, in previous works, the multi-task reasoning between ECPE and CE/EE is only achieved by implicit semantics-level interactions such as shared encoders.
For one thing, this is inconsistent with human intuition and the causal relations between ECPE and CE/EE, both of which are based on predictions or labels;
for another, the indicative information conveyed in semantics is implicit and relatively insufficient compared with prediction information.

On account of the above issues, we propose a new MTL framework based on Co-evolving Reasoning as shown in Fig.\ref{fig: framework}(b).
Firstly, in addition to the one-way message from CE/EE to ECPE, our MTL framework also models the reverse feedbacks from ECPE to CE/EE.
If ECPE predicts correctly, the indicative information transferred to CE/EE can improve them.
And the two improved subtasks can further promote ECPE reversely.
If ECPE predicts incorrectly, the defective information transferred to CE/EE can act as their feedback and make them rethink to provide better information for ECPE, which thus can further prompt CE/EE reversely.
Secondly, to achieve this virtuous cycle, we design the recurrent multi-task reasoning mechanism.
In this way, the knowledge of the three tasks can gradually evolve together and mutually prompt each other. 
Thirdly, we propose to exploit the correlations in predictions via introducing two kinds of prediction-level interactions: prediction-prediction and prediction-semantics interactions.
In this way, explicit correlative information conveyed by predictions (estimated label distributions) can flow in our MTL framework then facilitate \textit{Co-evolving Reasoning}.
And the semantics can get straightforward feedback at each step from the predictions then rethink to improve.

Furthermore, the MTL sequence structures employed in previous works are simply based on shared encoder and multi-task hidden states concatenations, which can hardly capture the causal relations.
This motivates us to seek a more effective method to sufficiently exploit the causalities among ECPE and CE/EE.
To this end, we design a novel multi-task relational graph (MRG), in which
there are three groups of nodes derived from the clauses in the document and corresponding to ECPE, CE, and EE, respectively.
Moreover, we design different relation types which correspond to the causal relations between ECPE and CE/EE.

To implement our MTL framework and conduct Co-evolving Reasoning on MRG, we propose a  Co-evolving Graph Reasoning Network (CGR-Net), whose core is a multi-task relational graph transformation (MRGT) cell.
CGR-Net first generates the task-specific hidden states and produces the initial estimated label distributions.
Then the MRGT cell recurrently takes the hidden states and label distributions of the three tasks as input and then updates them in three steps: projection of label distributions, relational local graph transformation, and non-local self-transformation.
Finally, the predictions of ECPE at the final step are used to extract the potential emotion-cause pairs.
And we design a harness loss based on logical constraints to force the three tasks to gradually promote each other in the virtuous cycle of Co-evolving Reasoning.

In summary, our contributions are three-fold:\\
(1) We propose a new MTL framework based on Co-evolving Reasoning and an MRG to exploit the correlations and causal relations sufficiently. To the best of our knowledge, our MTL framework is the first one allowing ECPE and CE/EE to promote each other recurrently, and MRG is the first MTL graph structure for ECPE.\\
(2) To implement our MTL framework and conduct Co-evolving Reasoning on MRG, we propose a novel CGR-Net, whose core is a multi-task relational graph transformation cell. \\
(3) Experimental results on the benchmark dataset show that our CGR-Net significantly outperforms existing state-of-the-art models. And further analysis proves the effectiveness of different components of CGR-Net and the superiority of MRG.

%% file: relatedwork.tex
\section{Related Works}\label{sec: relatedwork}
Emotion cause extractions (ECE) \cite{ece,coling2010,russo-etal-2011-emocause,neviarouskaya-aono-2013-ece,cicling,ecpcorpus,gui-ece-2017-question,conf/aaai/DingHZX19/ece,knowledgeece} is a long-standing task whose objective is to extract the causes of given emotion expressions in the document.
However, it requires that the emotions must be annotated manually, which constrains the practical application.
Therefore, recently \cite{ecpe} propose the ECPE task and a two-step solution 
while the error propagation may occur from the first step to the second.
To this end, recent works propose unified end-to-end models \cite{ecpe2d,rankcp,unifiedlabel,ecai,ding2020e2eecpe,ecai,localsearch,mtst,utos} to tackle ECPE in the MTL framework.

\cite{ecpe2d} propose a model integrating the 2D representation of emotion-cause pair, the interactions, and predictions.
\cite{rankcp} tackle ECPE from a ranking perspective and adopt kernel-based relative position embedding for ranking.
\cite{noveltagscheme} propose a tagging scheme coding the distance between the emotion clause and cause clause in an emotion-cause pair.
Based on this, \cite{mtst} propose a tag distribution refinement method that adjusts the output label distribution of ECPE using the ones of CE and EE according to a pre-defined rule.
However, the refinement method does not participate in model training, only working on the output of evaluation.

More recently, Multi-Granularity Semantic Aware Graph model (MGSAG) \cite{mgsag} incorporates fine-grained and coarse-grained semantic features jointly, aiming to resolve the distance limitation of clause semantics. And the Matrix Capsule-based multi-granularity framework (MaCa) \cite{maca} introduces the matrix capsule to obtain more fine-grained features of clause pairs, clustering the relationship of each clause pair.

Different from previous works, we (1) propose Co-evolving Reasoning, which allows the three tasks gradually and sufficiently promote each other; (2) introduce prediction-prediction and prediction-semantics interactions to model the explicit correlations and provide feedback for semantics which can rethink to improve; (3) effectively exploit the casual relations via designing a novel MRG.

%% file: method.tex
\section{Methodology}
Before delving into MRG and CGR-Net, we first introduce the task formulation in our work.

We cast ECPE as a \textit{tag} classification task and use the cause-centric tagging scheme \cite{noveltagscheme}.
Each clause $x_i$ has a two-tuples tag $y_i^t\!=\!(y_i^{t,c},y_i^{t,d})\!\in\!\mathcal{C}_t$
, where $y_i^{t,c}\!\in\!\{\text{C, O}\}$ denotes whether $x_i$ is a cause clause, and $y_i^{t,d}\!\in \!\{- \gamma, ..., -1, 0, 1, ..., \gamma,\perp \}$ denotes the distance between $x_i$ and its triggered emotion clause, while `$\perp$' always associates with `O', denoting that $x_i$ is a non-cause clause.
And $\gamma$ is a hyperparameter controlling the max span of emotion-cause pairs.
Thus ECPE (\textit{tag}) totally has $\left|\mathcal{C}_t\right|\!=\!2(\gamma\!+\!1\!)$ classes.

As for CE (\textit{cause}) and EE (\textit{emotion}), they are both formulated as binary classification tasks: $y_i^c\!\in\!\mathcal{C}_c\!=\!\{1,0\}$ and $y_i^e\!\in\! \mathcal{C}_e\!=\!\{1,0\}$.
\subsection{Constructing a MRG from a Document}
In this paper, we design a multi-task relational graph (MRG) $\mathcal{G}\!=\!(\mathcal{V},\mathcal{E},\mathcal{R})$ to exploit the causalities via modeling the self- and mutual-interactions of the three tasks (\textit{cause}, \textit{tag} and \textit{emotion}).
Each clause $x_i$ in document $\mathcal{D}$ derives three nodes $c_i$, $t_i$ and $e_i$, respectively for \textit{cause}, \textit{tag} and \textit{emotion}, thus $\left|\mathcal{V}\right|\!=\!3n$.
The edge $(i,j,r_{ij}\!)\in\!\mathcal{E}$ denotes the information propagation from node $i$ to node $j$, and $r_{ij}\!\in\! \mathcal{R}$ is the relation type of the edge.
Note that node $i$ and node $j$ may correspond to different tasks.
We define three kinds of rules to determine the connection between two nodes in MRG:

\textbf{Direction:} $(j,i,r_{j i})\! \in \!\mathcal{E}$ if $(i,j,r_{ij}) \!\in \!\mathcal{E}$.
In MRG, the information propagation between two nodes is bidirectional.
This guarantees the bidirectional correlations between \textit{tag} and \textit{cause}/\textit{emotion}.
\textbf{Local Connection:} $\forall (i,j,r_{ij}),\left|\operatorname{rdis}(i,j)\right|\!\leq\gamma$, where $\operatorname{rdis}(i,j)$ denotes the relative distance between the clauses of node $i$ and node $j$ in $\mathcal{D}$.
In general, the probability of two distant clauses having causal relation is relatively small regarding the cohesion and coherence of discourse \cite{discoursebook}. 
Therefore, the edges in MRG are based on local connections, and in this work, we constrain that the relative distance of two connected nodes' clauses in $\mathcal{D}$ ranges from $-\gamma$ to $\gamma$, consistent with the span range of the ECPE tag.

\begin{figure}[t]
 \centering
 \includegraphics[width = 0.6\textwidth]{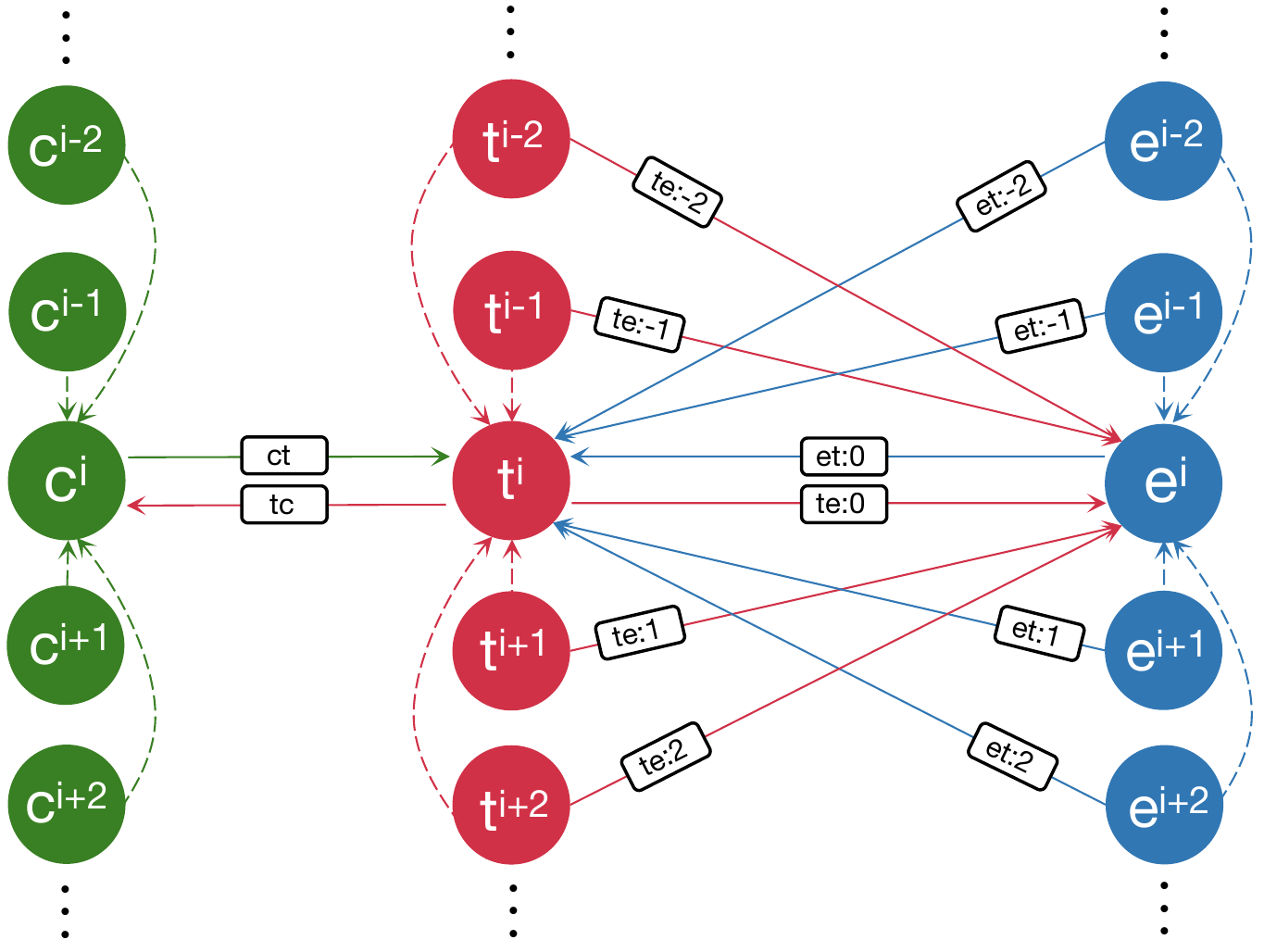}
 \caption{An example of MRG ($\gamma\!=\!2$). W.L.O.G, only the edges directed into $c_i, t_i$ and $e_i$ are illustrated. And self-loops are not shown for simplification. }
 \label{fig: mtgraph}
\end{figure}

\textbf{Relation Type for Causality:} Table \ref{table: mtgraph} lists the relation types in MRG. And an example of MRG is shown in Fig. \ref{fig: mtgraph}. 
\begin{table}[t]
\centering
\fontsize{8}{9}\selectfont
\caption{Relation types in MRG, w.l.o.g. $\gamma=2$. $\operatorname{I_t}(i)$ indicates node $i$ is a cause (c) node, tag (t) node or emotion (e) node. `-' denotes the set of [-2, -1, 0, 1, 2].}
\setlength{\tabcolsep}{2mm}{
\begin{tabular}{c|ccccccccccccccc}
\toprule
\rotatebox{90}{$r_{ij}$}                   &\rotatebox{90}{cc} &\rotatebox{90}{tt} &\rotatebox{90}{ee} &\rotatebox{90}{ct} &\rotatebox{90}{tc} &\rotatebox{90}{te:-2}  &\rotatebox{90}{te:-1} &\rotatebox{90}{te:0} &\rotatebox{90}{te:1} &\rotatebox{90}{te:2}&\rotatebox{90}{et:-2}  &\rotatebox{90}{et:-1} &\rotatebox{90}{et:0} &\rotatebox{90}{et:1} &\rotatebox{90}{et:2}   \\ \midrule
$\operatorname{I_t}(i)$    &c &t &e &c &t &t  &t  &t &t &t &e  &e  &e &e  &e     \\
$\operatorname{I_t}(j)$    &c &t &e &t &c &e  &e  &e &e &e &t  &t  &t  &t &e     \\
$\operatorname{rdis}(i, j)$&- &- &- &0 &0 &-2 &-1 &0 &1 &2 &-2 &-1 &0 &1 &2\\
\bottomrule
\end{tabular}}
\label{table: mtgraph}
\end{table}
To capture the self-task local contextual dependencies, we define $r_{ij}=cc$, $r_{ij}=tt$ and $r_{ij}=ee$ to model the local self-transformation of \textit{cause}, \textit{tag} and \textit{emotion}, respectively.
As for inter-task interactions, first of all, regarding the scheme of \textit{tag} task, there are two explicit causal relations between \textit{cause} and \textit{tag} tasks: (1) if $y_i^c=1$, then $y_i^{t,c}=\text{C}$, and vice versa; (2) if $y_i^c=0$, then $y_i^{t,c}=\text{O}$, and vice versa.
To model these causalities in MRG, we define $r_{ij}=ct$ and $r_{ij}=tc$ to achieve the mutual transformations between \textit{cause} and \textit{tag}.
Besides, there are four kinds of causal relations between \textit{tag} and \textit{emotion} tasks.
First, if $y_i^{t,c}=\text{C}$, there is at least one emotion clause among $x_{i-\gamma}\sim x_{i+\gamma}$:
$
y^e_{i-\gamma} = 1 \ \ \text{or}  \ \ y^e_{i-\gamma+1} = 1 \ \ \text{or} \ ... \ \text{or} \ \ y^e_{i+\gamma-1} = 1 \ \ \text{or} \ \ y^e_{i+\gamma}=1 
$.
Second, if there is no emotion clause among $x_{i-\gamma}\sim x_{i+\gamma}$:
$
y^e_{i-\gamma} = 0\ \  \text{and} \ \ y^e_{i-\gamma+1}=0 \ \  \text{and} \ ... \ \text{and} \ \  y^e_{i+\gamma-1}=0\ \  \text{and} \ \ y^e_{i+\gamma}=0
$,
then $y_i^{t,c}=\text{O}$ and $y_i^{t,d}=\perp$.
Third, if $y_i^{t,d}=m$, then $y^e_{i+m}=1$. Reversely, $y^e_{i+m}=1$ cannot deduce $y_i^{t,d}\!=\!m$, but intuitively if $p(y^e_{i+m}=1)$ increases, $p(y_i^{t,d}=m)$ should also increase.
Forth, if $y^e_{i+m}=0$, then $y_i^{t,d}\neq m$. Reversely, although $y_i^{t,d}\neq m$ cannot deduce $y^e_{i+m}=0$, intuitively if $p(y_i^{t,d}=m)$ decreases, $p(y^e_{i+m}=0)$ should increase.
To model these $tag$-$emotion$ causal relations in MRG, we define a set of relations represented by $r_{ij}=\operatorname{I_t}(i)\operatorname{I_t}(j):\operatorname{rdis}(ij)$, and some instances are shown in Fig. \ref{fig: mtgraph}.
For example, $r_{ij} = et:-2$ denotes the relation from node $i$ (an \textit{emotion} node) to node $j$ (a \textit{tag} node) and the relative distance between node $i$ and node$j$.

In MRG, each inter-task relation corresponds to a fine-grained relative distance, consistent with the definition of the tagging scheme.
Therefore, the inter-task graph transformations along these relations can achieve more sufficient and explicit multi-task reasoning.

\subsection{CGR-Net}
\begin{figure*}[t]
 \centering
 \includegraphics[width = \textwidth]{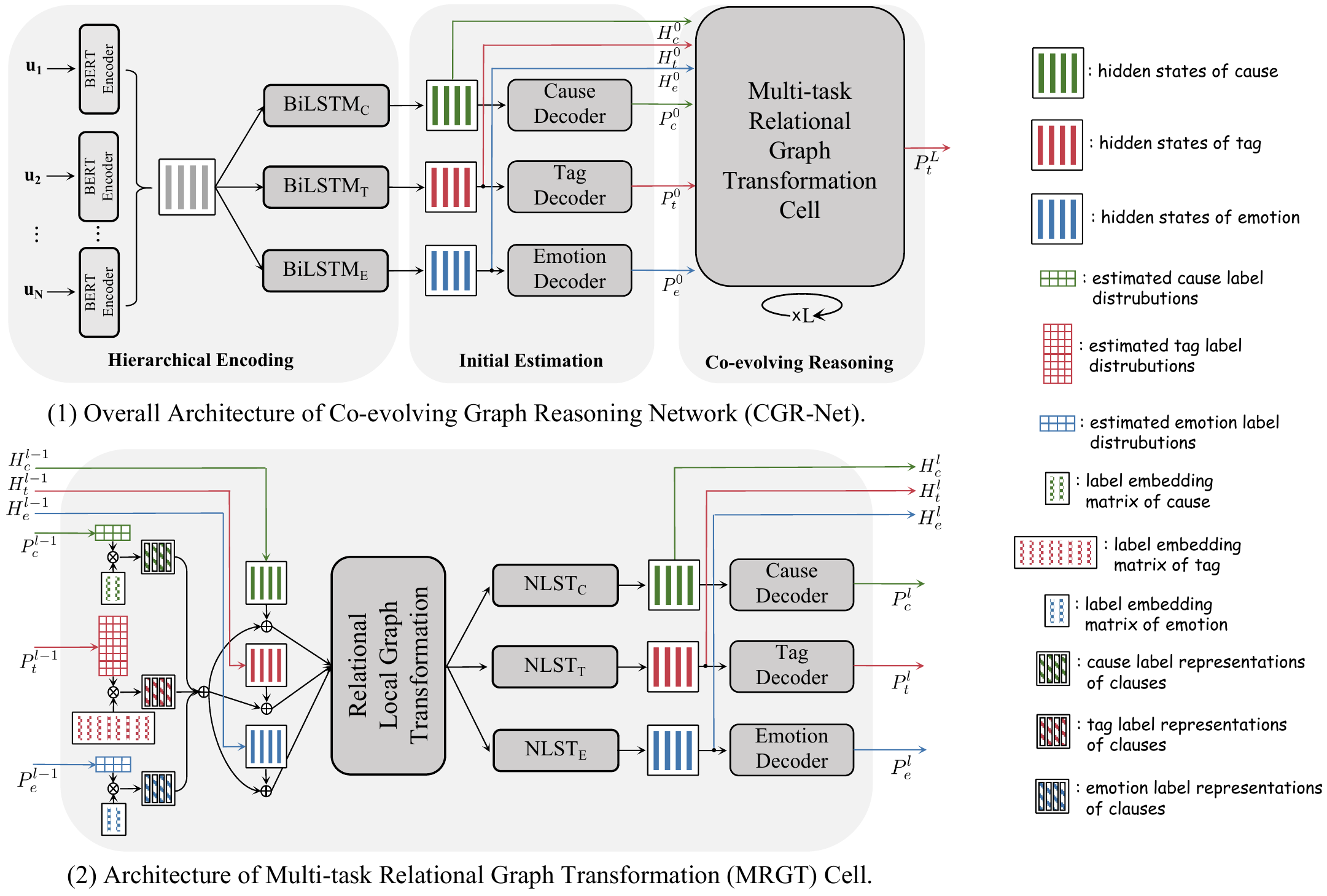}
 \caption{The architectures of CGR-Net and MRGT cell. NLST denotes Non-Local Self-Transformation.}
 \label{fig: model}
\end{figure*}
The overall architecture of our Co-evolving Graph Reasoning Network (CGR-Net) is shown in Fig. \ref{fig: model} (1).
It consists of three components: Hierarchical Encoding, Initial Estimation, and Co-evolving Reasoning. 
Next, we depict the procedures of these three components.
\subsubsection{Hierarchical Encoding}
\paragraph{Word-level Clause Encoding}
The objective of clause encoding is to generate a representation containing the word-level dependencies for each clause.
Following previous works, each clause is fed into BERT \cite{bert} encoder, then the last hidden state of \texttt{[CLS]} token is taken as the clause representation.
Now we obtain the sequence of clause representation for $\mathcal{D}$: $H=(h_0, ..., h_n)$.
\paragraph{Multi-task Clause-level Document Encoding}
In this paper, we utilize BiLSTM \cite{LSTM} to generate the context-sensitive clause hidden states via modeling the inter-clause dependencies.
To obtain task-specific clause hidden states for the three tasks, we separately apply three BiLSTMs over $H$ to obtain the initial clause hidden states for \textit{cause}, \textit{tag} and \textit{emotion}, respectively:
${H_c^0}=\left(h_{c,1}^0, ..., h^0_{c,n}\right)$, ${H_t^0}=\left(h_{t,1}^0, ..., h^0_{t,n}\right)$ and ${H_e^0}=\left(h_{e,1}^0, ..., h^0_{e,n}\right)$.

\subsubsection{Initial Estimation}
Since MRGT cell takes the three tasks' label distributions predicted in previous step as input, ${H_c^0}$, ${H_t^0}$ and ${H_e^0}$ are separately fed into Cause Decoder, Tag Decoder and Emotion Decoder to produce the initial estimated label distributions:
 \begin{equation}
 \begin{split}
  {P^0_{c}}&=  \{P^0_{c,i}\}_{i=1}^n, {P^0_{t}}=\{P^0_{t,i}\}_{i=1}^n, P^0_{e}=\{P^0_{e,i}\}_{i=1}^n\\ 
  P^0_{c,i} \!&= \!\operatorname{softmax}(\operatorname{MLP}_\text{c}(h^0_{c,i})) \!=\!\left[p^0_{c,i}[1], p^0_{c,i}[2]\right]\\
  P^0_{t,i} \!&=\! \operatorname{softmax}(\operatorname{MLP}_\text{t}(h^0_{t,i})) \!= \!\left[p^0_{t,i}[1], ..., p^0_{c,i}[\left|\mathcal{C}_t\right|]\right]\\
  P^0_{e,i} \!&=\! \operatorname{softmax}(\operatorname{MLP}_\text{e}(h^0_{e,i}))\!=\! \left[p^0_{e,i}[1], p^0_{e,i}[2]\right]\\
\end{split}
\end{equation}

\subsubsection{Co-evolving Reasoning}
Co-evolving Reasoning is achieved by the recurrent MRGT cell, whose details are shown in Fig. \ref{fig: model} (2).
At step $l$, MRGT cell takes two streams of inputs: 1) hidden states of the three tasks: ${H_c^{l-1}}\in\mathbb{R}^{n\times d}$, ${H_t^{l-1}\in\mathbb{R}^{n\times d}}$ and ${H_e^{l-1}\in\mathbb{R}^{n\times d}}$; 2) label distributions of the three tasks: $P^{l-1}_{c}$, $P^{l-1}_{t}$ and $P^{l-1}_{e}$. 
The procedure of an MRGT cell consists of three steps (1) projecting the input label distributions into vectors; (2) Relational Local Graph Transformation on MRG; (3) Non-local Self-Transformation.

\paragraph{Projection of Label Distribution}
To achieve the prediction-level interactions, the input label distributions should be projected into vector form, and thus they can participate in representation learning.
Accordingly, we ues $P_c^{l-1}$, $P_t^{l-1}$ and $P_e^{l-1}$ to respectively multiply the corresponding task-specific label embedding matrices $M^{e}_c\!\in\! \mathbb{R}^{\left|\mathcal{C}_c\right|\times d}$, $M^{e}_t\in\! \!\mathbb{R}^{\left|\mathcal{C}_t\right|\times d}$ and $M^{e}_e\in\! \!\mathbb{R}^{\left|\mathcal{C}_e\right|\times d}$, which are trained with the whole model.
Specifically, $x_i$'s label representations for the three tasks are obtained as:
 \begin{equation}
 e_{t,i}^l = \sum_{k=1}^{\left|\mathcal{C}_t\right|} p^{l-1}_{t,i}[k] \cdot v_t^{k}; \quad
e_{c,i}^l = \sum_{k'=1}^{\left|\mathcal{C}_c\right|} p^{l-1}_{c,i}[k'] \cdot v_c^{k'}; \quad
e_{e,i}^l = \sum_{k''=1}^{\left|\mathcal{C}_e\right|} p^{l-1}_{e,i}[k''] \cdot v_e^{k''}
\end{equation}
where $v_t^k$, $v_c^{k'}$ and $v_e^{k''}$ denotes the label embeddings of \textit{tag}, \textit{cau} and \textit{emo}, respectively. 

\paragraph{Relational Local Graph Transformation}
Since MRG is based on local connections, we conduct relational local graph transformation inspired from \cite{rgcn,kagrmn,jair,dignet,darer,coguiding,relanet} for multi-task reasoning.
To achieve the self- and mutual-interactions between the semantics and predictions of the three tasks, for each node in MRG, we superimpose its corresponding clause's label representations of the three tasks on its hidden state:
 \begin{equation}
 \begin{split}
 e_i^l =& e_{c,i}^l + e_{t,i}^l + e_{e,i}^l,\\
\hat{h}_{c,i}^l = h_{c,i}^{l-1}\! +\!  e_i^l; \quad \hat{h}_{t,i}^l =& h_{t,i}^{l-1}\! + \! e_i^l; \quad \hat{h}_{e,i}^l = h_{e,i}^{l-1}\! + \! e_i^l
\end{split}
\end{equation}
Thus each node representation contains the task-specific semantic features as well as the explicit correlative information conveyed by label representations, which are then integrated together into the relational local graph transformation to achieve semantics-level and prediction-level interactions.

Specifically, the relational local graph transformation updates the nodes on MRG as follows:
\begin{equation}
 \overline{h}_{i}^{l}=W_{1} \hat{h}_{i}^t + \sum_{r \in \mathcal{R}} \sum_{j \in \mathcal{N}_{i}^{r}} \frac{1}{\left|\mathcal{N}_{i}^{r} \right|} W^r_{2} \hat{h}_{j}^l
\end{equation}
where $W_{1}$ is the self-message matrix and $W_2^r$ is the relation-specific matrix.
$\mathcal{N}^r_i$ denotes the neighbors set of node $i$ along corresponding to the relation $r$.
Now we obtain the updated 
hidden states: 
$\overline{H}_c^l$, $\overline{H}_t^l$ and $\overline{H}_e^l$.
\paragraph{Non-Local Self-Transformation}
Despite the advantages of the relational local graph transformation, it has two potential issues: 
(1) due to the local self-transformation, some beneficial contextual dependencies between a node and its distant same-task nodes may be lost; 
(2) the information fusion weaken the task-specificity of the nodes to some extent, which is against predictions.
To this end, inspired by \cite{aalstm,ijcaisubgraph}, we conduct non-local self-transformation (NLST) over the sequence of nodes of each task, and this is implemented by a task-specific BiLSTM which can capture long-range dependencies.
The final hidden states of the three tasks at step $l$ are obtained by:
\begin{equation}
H_c^l=\text{NLST}_\text{C}(\overline{H}_c^l); \quad
H_t^l=\text{NLST}_\text{T}(\overline{H}_t^l); \quad
H_e^l=\text{NLST}_\text{E}(\overline{H}_e^l)
\end{equation}
Then $H_c^l$, $H_t^l$ and $H_e^l$ are fed to respective decoders to produce $P_c^l$, $P_t^l$ and $P_e^l$.  

\subsubsection{Optimization with Logical Constraints}
In CGR-Net, there are two vital logic rules.
First, the label distributions estimated in the previous step should be relatively good to provide effective label representations for the current step.
Otherwise, much incorrect and misleading explicit correlations would be introduced, harming multi-task reasoning.
Second, ECPE and CE/EE are supposed to gradually promote each other via
capturing more and more beneficial mutual knowledge and correlations in Co-evolving Reasoning.
In other words, the estimated label distributions should be gradually improved along the steps.
To satisfy these two rules, we propose a harness loss $\mathcal{L}_{harn}$ that includes two terms: estimate loss $\mathcal{L}_{est}$ and margin loss $\mathcal{L}_{marg}$, corresponding to the two rules, respectively.
\paragraph{Estimate Loss}
Formally, $\mathcal{L}_{est}$ is the cross-entropy loss.
For ECPE task, $\mathcal{L}^{tag,l}_{est}$ is defined as:
\begin{equation}
 \mathcal{L}^{tag,l}_{est}=\frac{1}{n}\!\sum_{i=1}^{n}\sum_{k=1}^{\left|\mathcal{C}_t\right|}y_{t,i}^k \text{log}\left(p_{t,i}^{l}[k]\right),\label{eq: estloss}
\end{equation}

\paragraph{Margin Loss}
$\mathcal{L}_{marg}$ works on the label distributions output in two adjacent steps, forcing CGR-Net to produce better predictions at step $l$ than step $l-1$.
For ECPE task, $\mathcal{L}^{tag,l}_{marg}$ is defined as:
\begin{equation}
 \mathcal{L}^{tag,(l,l-1)}_{marg}=\frac{1}{n}\sum_{i=1}^{n}\sum_{k=1}^{\left|\mathcal{C}_t\right|}y_{t,i}^k \ \text{max}(0,p_{t,i}^{l-1}[k]-p_{s,i}^{l}[k])\label{eq: marginloss}
\end{equation}
\paragraph{Harness loss}
$\mathcal{L}_{harn}$ is the weighted sum of $\mathcal{L}_{est}$ and $\mathcal{L}_{marg}$.
For ECPE task, $\mathcal{L}^{tag}_{harn}$ is defined as:
\begin{equation}
 \mathcal{L}^{tag}_{harn}=\sum_{l=0}^{L-1}\mathcal{L}^{tag,l}_{est} + \beta * \sum_{l=1}^{L}\mathcal{L}^{tag,(l,l-1)}_{marg}\label{eq: harnessloss}
\end{equation}
where $\beta$ is a hyper-parameter balancing the impact of the two kinds of punishments.
\paragraph{Final Training Objective}
The total loss for ECPE task ($\mathcal{L}_{tag}$) is the sum of $\mathcal{L}^{tag}_{harn}$ and $\mathcal{L}^{tag}_{pred}$:
\begin{equation}
 \mathcal{L}_{tag} =\mathcal{L}^{tag}_{pred} + \mathcal{L}^{tag}_{harn} \label{eq: sentiloss}
\end{equation}
where $\mathcal{L}^{tag}_{pred}$ is the cross-entropy loss of the produced tag label distributions at the final step $L$:
\begin{equation}
 \mathcal{L}^{tag}_{pred}=\frac{1}{n}\sum_{i=1}^{n}\sum_{k=1}^{\left|\mathcal{C}_t\right|}y_{t,i}^k \ \text{log}\left(p_{t,i}^{L}[k]\right) \label{eq: predloss}
\end{equation}

The total losses of CE ($\mathcal{L}_{cau}$) and EE ($\mathcal{L}_{emo}$) can be derivated like  \cref{eq: estloss,eq: marginloss,eq: harnessloss,eq: sentiloss,eq: predloss}.

The final training objective of CGR-Net is the weighted sum of the total losses of the three tasks:
\begin{equation}
 \mathcal{L} = \alpha * \mathcal{L}_{tag}+ \frac{1-\alpha}{2} \mathcal{L}_{cau} + \frac{1-\alpha}{2}\mathcal{L}_{emo} 
\end{equation}
where $\alpha$ is a hyperparameter balancing the three tasks and it is intuitively set as 0.5 in this work.

%% file: experiment.tex
\section{Experiments}
\subsection{Datasets and Evaluation Metrics}
The only benchmark dataset for ECPE task is released by \cite{ecpe} who construct it on an emotion-cause extraction corpus \cite{ecpcorpus}.
The dataset totally consists of 1,945 documents, among which 1,746 ones have one emotion-cause pair, 177 ones have two emotion-cause pairs, and 22 ones have more than two emotion-cause pairs.
The average number of clauses per document is 14.77, and the max number is 73.

Following previous works, we adopt the 10-fold cross-validation for evaluations.
And the averages of precision (P), recall (R), and F1-score over ten runs are adopted as metrics.
Besides ECPE, we also report the results of EE and CE, which are evaluated based on the emotion clauses and cause clauses in the extracted emotion-cause pairs.

\subsection{Implement Details} \label{sec: implement}
 We adopt the BERT$_{Chinese}$ implemented in PyTorch \cite{transformers} as the clause encoder.
 And the three decoders are implemented as three 2-layer MLPs whose hidden size is set as 256.
The AdamW optimizer \cite{adamw} is used for model training, and the learning rate is $1e^{-5}$ for BERT and $1e^{-4}$ for other modules.
The dimension $d$ is 512, 
the max span $\gamma$ is 3 and the margin loss coefficient $\beta$ is $1e^{-3}$.
The step number of Co-evolving Reasoning is 3.
The dropout rate is 0.1, and the batch size is 4.
The epoch number is 10, and the early stopping strategy is adopted.
All experiments are conducted on a DGX A100 server.

\begin{table*}[t]
\centering
\fontsize{11}{13}\selectfont
\caption{Results comparison on ECPE task and the two subtasks. All scores are averages over 10 runs. All models adopt BERT for clause encoding. $^\natural$ denotes the results are reproduced by us.
$^\dag$ and $^\ddag$ denote the results are retrieved from \cite{mtst} and \cite{utos}, respectively. $^*$ denotes our CGR-Net significantly overpasses M$_{10}$ and M$_{11}$  with $p<0.05$ under t-test. }
\setlength{\tabcolsep}{1mm}{
\resizebox{\textwidth}{!}{%
\begin{tabular}{c|ccc|ccc|ccc}
\toprule
\multirow{2}{*}{Models} & \multicolumn{3}{c|}{Emotion-Cause Pair Ext.}  &  \multicolumn{3}{c|}{Emotion Ext.} & \multicolumn{3}{c}{Cause Ext.}\\ 
                & P (\%)   & R (\%)  & F1 (\%)    & P (\%)   & R (\%)  & F1 (\%)  & P (\%)   & R (\%)  & F1 (\%) \\\midrule
M$_1$: ECPE-2D(BERT)         &72.92 &65.44 &68.89 &86.27 &92.21 &\textbf{89.10} &73.36 &69.34 &71.23 \\
M$_2$:  Hier-BiLSTM-BERT       &75.37 &64.34 &69.26  &88.80 &74.70 &81.00 &78.03 &65.35 &70.96 \\
M$_3$:  PairGCN-BERT           &76.92 &67.91 &72.02 &88.57 &79.58 &83.75 &79.07 &69.28 &73.75 \\
M$_4$:  TransECPE$^\dag$      &77.08 &65.32&70.72 &88.79 &83.15&85.88&78.74 &66.89&72.33\\
M$_5$:  RankCP+BERT$^\ddag$       &68.21 &74.83 &71.21  &86.79 &89.26 &87.97 &72.62 &76.46 &74.37\\
M$_6$:  UTOS+BERT  &73.89 &70.62 &72.03  &88.15 &83.21 &85.56 &76.71 &73.20 &74.71 \\
 M$_7$: ECPE-MLL(BERT)& 77.00 &72.35 &74.52 &86.08 &91.91 & 88.86 & 73.82 &79.12 & 76.30\\
 M$_8$: MGSAG(BERT)& 77.43 &73.21 &75.21 &92.08 &92.11 & 87.17 & 79.79 &74.68 & 77.12\\
 M$_9$: MaCa(BERT)& 80.47 &72.15 &73.87 &88.19 &89.55 & 87.04 & 78.41 & 72.60& 74.35\\
  \hline
M$_{10}$:  SLNT+BERT$^\natural$          &73.56 & 68.57  &70.85 &84.77  &80.61 &82.51 &75.94 &70.99 &73.25 \\
M$_{11}$:  MTST+Refinement$^\natural$     & 77.14 &67.81 & 72.11  &88.25  &79.01  &83.31& 79.18 &69.79 &74.12 \\ \midrule
CGR-Net (ours)   &77.62 &\textbf{75.49}$^*$ & \textbf{76.48}$^*$ &89.65 &86.23$^*$ &87.75$^*$ & 79.68 & 77.84$^*$ &\textbf{78.75}$^*$ \\
 \bottomrule
\end{tabular}}}
\label{table: results}
\end{table*}

\subsection{Compared Baselines}
We compare our CGR-Net with the following two groups of baselines.\\

\textbf{Group 1}:
$\text{M}_1$: ECPE-2D (BERT) \cite{ecpe2d};
$\text{M}_2$ :Hier-BiLSTM-BERT; $\text{M}_3$: PairGCN-BERT \cite{pairgcn};
$\text{M}_4$: TransECPE \cite{transecpe};
 $\text{M}_5$: RankCP+BERT \cite{rankcp};
$\text{M}_6$: UTOS+BERT \cite{utos};
$\text{M}_7$: ECPE-MLL(BERT) \cite{ding2020e2eecpe}; 
$\text{M}_8$: MGSAG(BERT) \cite{mgsag}; $\text{M}_9$: MaCa(BERT) \cite{maca};

\textbf{Group 2}:
$\text{M}_8$: SLNT + BERT \cite{noveltagscheme};
$\text{M}_9$: MTST + Refinement \cite{mtst}.

Our CGR-Net uses the same ECPE tagging scheme with the second group of baselines.

\subsection{Main Results} \label{sec: mainresult}

The overall results on ECPE and the two subtasks are shown in Table \ref{table: results}. 
We can observe that CE is much harder than EE, and CE determines the result of ECPE to a large extent. The reason is that CE plays a key role in identifying the causalities, which is difficult for machines. Our CGR-Net significantly outperforms the previous best-performing model M8 by 1.7\%, 0.7\%, and 2.1\% in terms of F1 on ECPE, EE and CE, respectively. Using the same tagging scheme, our CGR-Net overpasses M$_{10}$ and M$_{11}$ by 6.1\%, 5.3\%, and 6.2\% in F1 on the three tasks. In particular, we can find that the superior F1 of CGR-Net comes from the high recall and competitive precision. In contrast, previous models generally obtain low recalls because their single-round multi-task reasoning process and the one-way message only from CE/EE to ECPE are not competent enough to discover the emotion cause pairs sufficiently. Moreover, their implicit semantics interactions cannot effectively exploit the causal relations. We can find that our CGR-Net overpasses M$_{10}$ and M$_{11}$ by 10.1\%, 7.0\%, and 9.6\% in terms of recall on the three tasks. This demonstrates that through Co-evolving Reasoning, CGR-Net can discover much more ground-truth emotion-cause pairs than baselines, while there are not many wrong-extracted pairs at the same time. CGR-Net’s satisfying results come from the advantages of MRG, the well-designed supervision signals, and the advanced architecture of MRGT.

\subsection{Variants of MRG Structure}
\begin{table}[t]
\centering
\fontsize{8}{10}\selectfont
\caption{Results on different variants of MRG.}
\setlength{\tabcolsep}{2mm}{
\begin{tabular}{cccc}
\toprule
\multirow{2}{*}{Variants} &ECPE  &EE   &CE    \\ \cline{2-4}
& F1 (\%)  & F1 (\%)  & F1 (\%) \\\midrule
MRG($\gamma=3$) &76.48  &87.75   &78.75\\\midrule
MRG($\gamma=1$)  & 75.15 ($\downarrow$ 1.33) &86.31 ($\downarrow$ 1.44)   &77.32 ($\downarrow$ 1.43)\\
MRG($\gamma=2$)  & 75.49 ($\downarrow$ 0.99) &87.55 ($\downarrow$ 0.20)   &77.79 ($\downarrow$ 0.96)\\
MRG($\gamma=4$)  & 74.87 ($\downarrow$ 1.61) & 86.78 ($\downarrow$ 0.97)   &77.21 ($\downarrow$ 1.54)\\ \hline
OWM              & 74.71 ($\downarrow$ 1.77 & 86.30 ($\downarrow$ 1.45)   &77.22($\downarrow$ 1.53)) \\
NoRel            &74.63 ($\downarrow$ 1.85) &85.33 ($\downarrow$ 2.42) &77.10 ($\downarrow$ 1.65)  \\
FCG              &73.65 ($\downarrow$ 2.83)  &84.89 ($\downarrow$ 2.86) &76.56 ($\downarrow$ 2.19)  \\
\bottomrule
\end{tabular}
}
\label{table: graph_variants}
\end{table}

In this section, we investigate how the structure of MRG would affect our CGR-Net's performance by applying different structures to MRG.
Except for the original MRG($\gamma=3$), we design four variants: 
(1) MRG($\gamma=1$), MRG($\gamma=2$) and MRG($\gamma=4$), which have different span limitation; 
(2) OWM (one-way message), in which the edges from \textit{tag} nodes to \textit{emotion} nodes and \textit{cause} nodes are deleted, thus there is only one-way message from EE/CE to ECPE, like previous works; 
(3) NoRel, in which there is no relation on edges; 
(4) FCG (fully-connected graph), in which all \textit{tag} nodes are fully connected with all \textit{emotion} nodes and all \textit{cause} nodes, while there is no relation.
The results over these variants are listed in Table \ref{table: graph_variants}.

Several instructive observations can be made from the results.
\textbf{Firstly}, with $\gamma$ varying from 1 to 4, the results first increase and then decrease.
The reason is that too small $\gamma$ cannot capture all emotion-pairs, while too large $\gamma$ makes it much harder to predict correctly because $\left|\mathcal{C}_t\right|$ is directly proportional to $\gamma$.
More than 95\% emotion-cause pairs' spans in the dataset do not exceed 3, so intuitively $\gamma=3$ performs best, and the results prove this.
And this is consistent with the report of \cite{mtst}.
\textbf{Secondly}, compared with the original MRG, OWM's performances on all tasks drop significantly.
This proves that the reverse feedbacks from ECPE to EE/CE is crucial, while previous works ignore them. In this paper, we propose a new MTL framework based on the novel Co-evolving Reasoning mechanism to solve this issue.
\textbf{Thirdly}, without relations, NoRel performs much worse than the original MRG.
The distinct decrease of results proves that our designed relations are indispensable for capturing the casualties between ECPE and EE/CE and significantly improve the performances.
\textbf{Finally}, FCG performs even worse than NoRel.
This is because causalities often exist between close clauses.
In FCG, useless information from distant nodes is integrated into the current node, making the crucial information diluted and discarded, resulting in poor results.
And this proves the validity of the local connection rule in MRG.

\subsection{Investigation of Supervision Signals}
\begin{table}[t]
\centering
\fontsize{8}{10}\selectfont
\caption{Results of removing different loss terms.}
\setlength{\tabcolsep}{2mm}{
\begin{tabular}{cccc}
\toprule
\multirow{2}{*}{Variants}  &EE   &CE  &ECPE  \\ \cline{2-4}
& F1 (\%)  & F1 (\%)  & F1 (\%) \\\midrule
CGR-Net  &76.48   &87.75& 78.75\\\midrule
$-\mathcal{L}_{emo}$ & 74.51 ($\downarrow$ 1.97) &86.21 ($\downarrow$ 1.54)&77.03 ($\downarrow$ 1.72) \\
$-\mathcal{L}_{cau}$ &74.28 ($\downarrow$ 2.20)  &86.25($\downarrow$ 1.50) &76.94 ($\downarrow$ 1.81) \\
$-\mathcal{L}_{emo}-\mathcal{L}_{cau}$  & 73.76 ($\downarrow$ 2.72)  &85.05 ($\downarrow$ 2.70) &76.13 ($\downarrow$ 2.62)  \\ \hline
$-\mathcal{L}_{est}$    &75.43 ($\downarrow$ 1.05)  &87.54 ($\downarrow$ 0.21) &77.97($\downarrow$ 0.78)\\
$-\mathcal{L}_{marg}$  &74.22 ($\downarrow$ 2.26) &86.51($\downarrow$ 1.24) &76.64 ($\downarrow$ 2.11) \\
$-\mathcal{L}_{harn}$ &74.89 ($\downarrow$ 1.59)  &85.61 ($\downarrow$ 2.14) &77.60 ($\downarrow$ 1.15)  \\
\bottomrule
\end{tabular}
}
\label{table: loss_ablation}
\end{table}

To investigate the necessities of different supervision signals, we remove different loss terms, and the results are listed in Table \ref{table: loss_ablation}.
Firstly, we can observe that removing $\mathcal{L}_{emo}$ or $\mathcal{L}_{cau}$ both lead to obvious result decreases on all tasks.
This is because our model exploits the beneficial mutual correlations between ECPE and the two subtasks.
So removing the supervision signal of a subtask harms not only the performance of itself but also the performances of ECPE and another subtask.
If both subtasks do not have supervision signals, the performances of all tasks drop dramatically.
Then we can find that the performances decrease remarkably if $\mathcal{L}_{harn}$ or any of its two terms is removed.
This is because without $\mathcal{L}_{harn}$ CGR-Net is hard to achieve the virtuous cycle of Co-evolving Reasoning.
\subsection{Ablation Study of MRGT Cell}
\begin{figure*}[t]
 \centering
 \includegraphics[width = 0.55\textwidth]{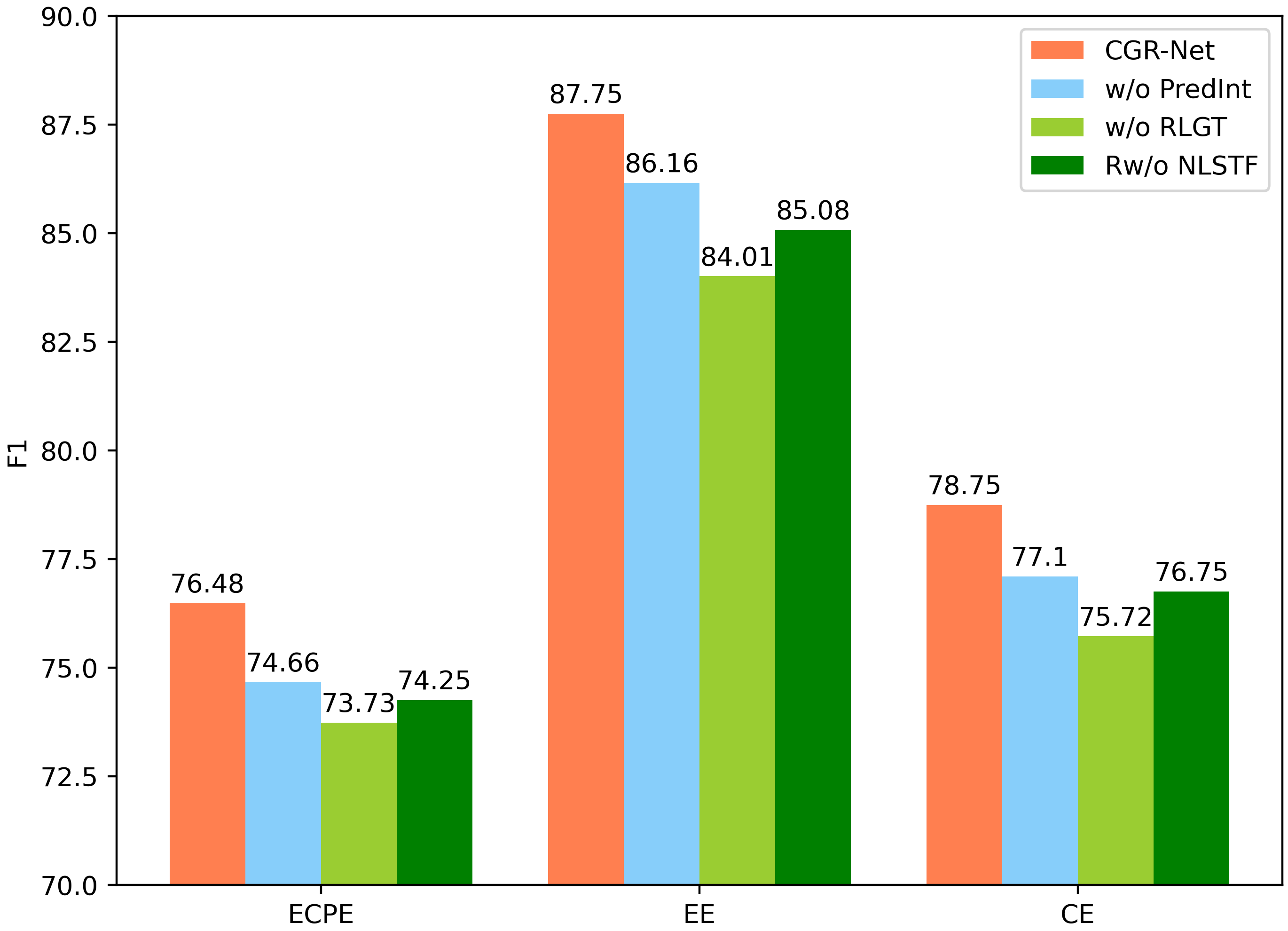}
 \caption{Ablation results of MRGT cell.}
 \label{fig: cell ablat}
\end{figure*}

We conduct ablation experiments to study the efficacies of the components in MRGT cell, and the results are listed in Fig. \ref{fig: cell ablat}. When removing prediction-level interactions (PredInt), the performances drop significantly. This proves that only relying on semantics-level interactions is insufficient for multi-task reasoning. An essential advantage of our model is achieving prediction-level interactions that convey explicit correlations and provide feedback for semantics that can then rethink to improve. Without RLGT, Co-evolving Reasoning cannot be achieved, causing the worst results. Removing NLST also leads to sharp decreases in results. This is because without NLST the hidden states cannot obtain long-range crucial contextual information, and the three streams of hidden states output at each step are not task-specific enough, which both harm the predictions.

\begin{figure}[t]
 \centering
 \includegraphics[width = 0.55\textwidth]{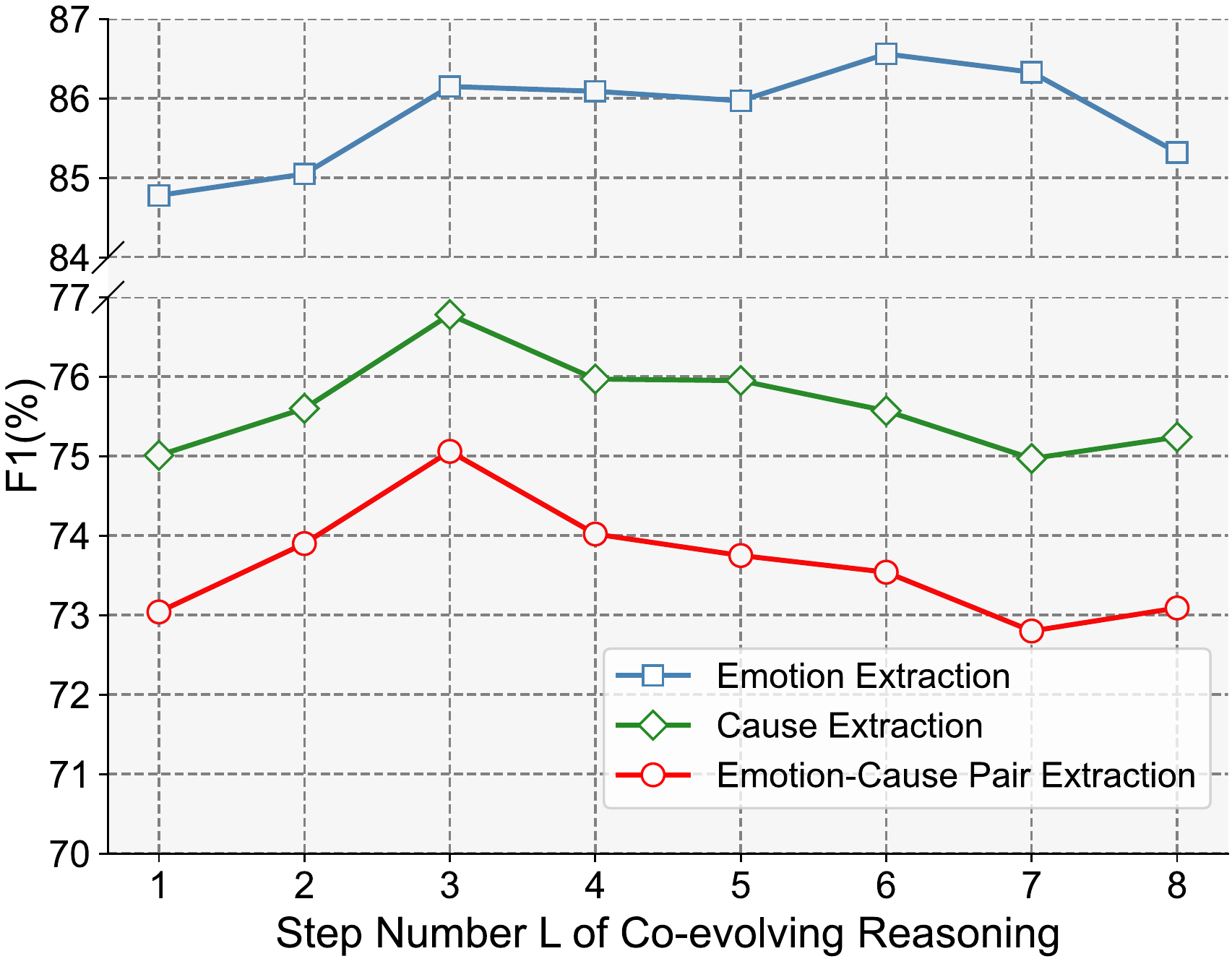}
 \caption{CGR-Net's performances on different $L$. }
 \label{fig: step}
\end{figure}
\subsection{Step Number of Co-evolving Reasoning}
We plot the performance (F1) trends of CGR-Net on the three tasks over different Co-evolving Reasoning step numbers, as presented in Fig. \ref{fig: step}.
The best overall performances are achieved when $L= 3$, which justifies the step number setting in Sec. \ref{sec: implement}.
Generally, the performances of ECPE and the two subtasks steadily increase until $L=3$, while then having dropping trends or fluctuate in a relatively narrow range when $L$ continues increasing.
This indicates that ECPE and CE/EE can gradually promote each other in the process of Co-evolving Reasoning, whose advantage is validated.
However, after the performance reaches its peak, more steps lead to decreasing.
We speculate the possible reason is that too many Co-evolving Reasoning steps may cause redundant information and over-fitting.

%% file: conclusion.tex
\section{Conclusion and Prospect}\label{sec: conclusion}
In this paper, we improve ECPE on three aspects.
First, we propose a new MTL based on Co-evolving Reasoning, allowing ECPE and its two subtasks to promote each other gradually.
Besides, prediction-level interactions are integrated to model the explicit correlations. 
Second, we design a novel multi-task relational graph (MRG) to sufficiently exploit the causal relations. 
Finally, we propose a Co-evolving Graph Reasoning Network (CGR-Net) to implement our framework and conduct Co-evolving Reasoning on MRG.
Experiment results demonstrate the superiority of our method, and detailed analyses further validate the advantages.

This work contributes a new paradigm not only for ECPE but also for a group of scenarios in which different tasks share the same input sequence.
Future works include improving our method on ECPE and applying our paradigm to other MTL tasks.